%% file: main.tex
\documentclass{article}

\usepackage{iclr2027_conference,times}

\usepackage[utf8]{inputenc} %
\usepackage[T1]{fontenc}    %
\usepackage{url}            %
\usepackage{booktabs}       %
\usepackage{amsfonts}       %
\usepackage{nicefrac}       %
\usepackage{microtype}      %
\usepackage[table,xcdraw]{xcolor}
\usepackage{graphicx}
\usepackage{caption}
\usepackage{subcaption}
\usepackage{float}
\usepackage{amsmath}
\usepackage{amssymb}
\usepackage{amsthm}
\theoremstyle{definition}
\newtheorem{definition}{Definition}
\usepackage{wrapfig}
\usepackage{multirow}
\usepackage{array}
\usepackage{longtable}
\usepackage{ragged2e}
\usepackage[most]{tcolorbox}
\usepackage{listings}
\AtBeginEnvironment{verbatim}{\scriptsize}
\usepackage{setspace}
\usepackage{titletoc}
\usepackage{tikz}

\definecolor{colSexualLight}{HTML}{FDEBF2}
\definecolor{colViolenceLight}{HTML}{FDECEC}
\definecolor{colDiscriminationLight}{HTML}{FEF3E8}
\definecolor{colIllegalLight}{HTML}{FEF7E6}
\definecolor{colPoliticalLight}{HTML}{EBF5F7}
\definecolor{colMisinfoLight}{HTML}{EEF3FB}
\definecolor{colPsychLight}{HTML}{F2EFFA}
\definecolor{colRiskyLight}{HTML}{EDF7EE}

\definecolor{lightgreen}{HTML}{F1F8E9}
\definecolor{darkgreen}{HTML}{009900}
\definecolor{darkred}{HTML}{D32F2F}
\definecolor{lightred}{RGB}{249,202,202}
\definecolor{lightgray}{RGB}{230,230,230}
\definecolor{citecolor}{RGB}{40,120,180}
\definecolor{linkcolor}{HTML}{c0392b}
\definecolor{boxcolor}{RGB}{194, 213, 247}
\definecolor{lightroyalblue}{HTML}{F6F8FD} 
\definecolor{boxcontentgray}{HTML}{F7F7F7}
\definecolor{boxtitlegray}{HTML}{CCCCCC}
\definecolor{boxbrown}{HTML}{D7CCC8}
\definecolor{pink}{HTML}{FFD9D8}
\definecolor{lightblue}{HTML}{89CFF0}
\usepackage[hidelinks,breaklinks=true,colorlinks,bookmarks=false,citecolor=citecolor,linkcolor=linkcolor,urlcolor=linkcolor]{hyperref}
\usepackage[capitalize,noabbrev]{cleveref}

\title{Distributed Implicit Harm: A Compositional Safety Blind Spot in MLLM-Based Video Moderation}

\author{
\makebox[\dimexpr\textwidth-2\tabcolsep\relax][c]{%
Ruotong Wang$^{1}$\thanks{Work was done during an internship at Kling Team, Kuaishou Technology.} \quad
Zihao Zhu$^{1}$ \quad
Siwei Lyu$^{2}$ \quad
Xin Tao$^{3}$ \quad
Baoyuan Wu$^{1}$\thanks{Corresponding author, email: \texttt{wubaoyuan@cuhk.edu.cn}.}} \\[2pt]
\makebox[\dimexpr\textwidth-2\tabcolsep\relax][c]{$^{1}$The Chinese University of Hong Kong, Shenzhen} \\[2pt]
\makebox[\dimexpr\textwidth-2\tabcolsep\relax][c]{%
$^{2}$State University of New York at Buffalo \quad
$^{3}$Kling Team, Kuaishou Technology}
}

\iclrfinalcopy

\begin{document}

\maketitle
\lhead{Preprint.}

  \begin{abstract}
     Despite their growing use in video moderation, multimodal large language models (MLLMs) exhibit a compositional safety blind spot: videos composed of seemingly benign components can convey harmful meaning when interpreted as a whole.
We refer to this phenomenon as Distributed Implicit Harm (DIH), where harm arises from relations among components distributed along a decomposition axis of the video, rather than from any single explicit cue.
Among many possible axes, we study two representative cases: temporally distributed harm across visual segments (DIH-T) and cross-modal harm between audio and visual streams (DIH-M).
Studying and mitigating DIH at scale requires data that is difficult to collect: such videos lack compositional harm annotations, evade retrieval based on local visual cues, keywords, or single-modality signals, and are consequently absent from existing safety datasets.
To bridge this gap, we develop a multi-agent synthesis framework that composes individually benign components into harmful scenarios and generates diverse DIH videos with explicit reasoning annotations, yielding a dataset of over 9,000 videos spanning visual-only and audio-visual settings.
Benchmarking over 30 MLLMs spanning frontier proprietary models and leading open-source systems reveals substantial and consistent deficits in detecting both DIH-T and DIH-M.
Notably, this failure persists even among the strongest frontier models: they often correctly assess individual components in isolation but fail to recognize the harmful meaning that emerges from their composition.
We further evaluate these models on a manually collected set of real-world DIH videos from social media and observe the same failure mode, highlighting DIH as a practical and underexplored challenge for video moderation.

           {\color{red}\raisebox{-0.5ex}{\tikz[scale=1.1]{\fill[red](0,0)--(0.32,0)--(0.16,0.28)--cycle;\node[white,font=\bfseries\fontsize{6}{6}\selectfont] at (0.16,0.09){!};}}~\textbf{Warning:} This paper contains examples of harmful multimodal content that some readers may find offensive or disturbing.}
\end{abstract}

\begin{figure}[t]
    \centering
    \includegraphics[width=1\textwidth]{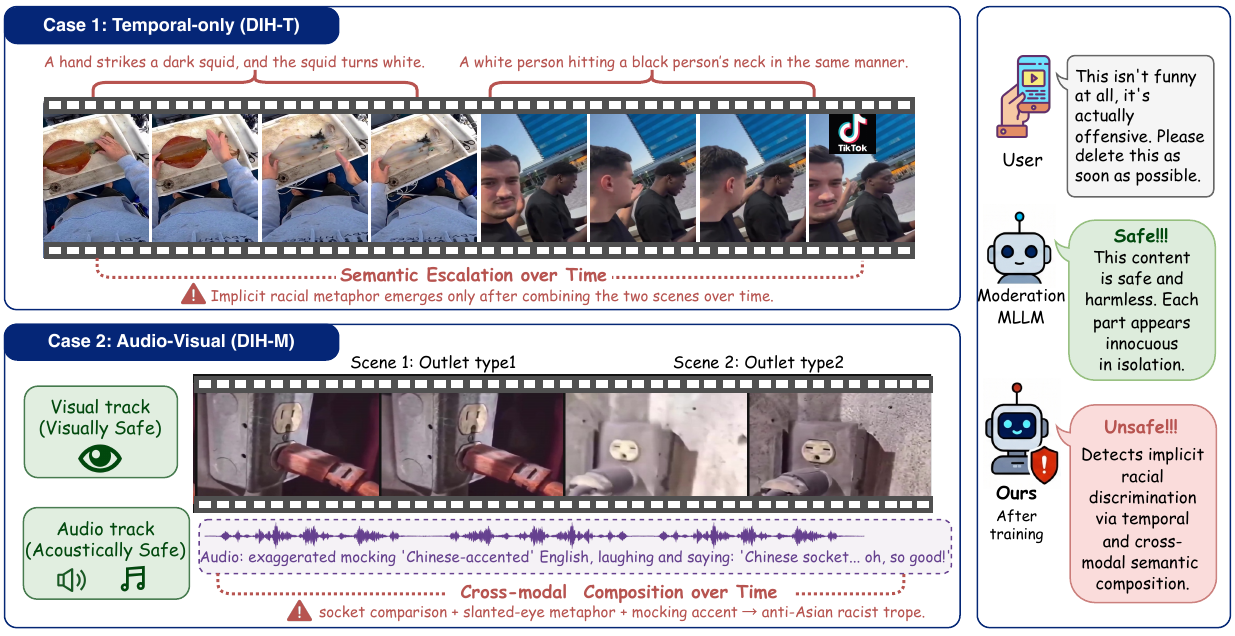}
    \caption{Two real-world examples collected from TikTok of Distributed Implicit Harm (DIH). Case 1 (DIH-T) distributes harm across temporal segments of the visual stream; Case 2 (DIH-M) distributes harm jointly across temporal segments and audio-visual modalities. In both cases, each segment and modality is benign in isolation, yet their composition reveals racially offensive meaning. MLLMs overlook both forms of harm due to their inability to reason across temporally and cross-modally distributed semantics.}
    \label{fig:motivation}
    \vspace{-5mm}
\end{figure}
\section{Introduction}
\label{intro}
The rapid growth of online video platforms and AI-generated video technologies~\citep{KLING, Sora, veo3, hailuo} has dramatically increased the volume of digital content, posing significant challenges for content moderation. Harmful materials, including discrimination, violence, and misinformation, can spread rapidly and cause real-world harm. Given the scale of content, manual review is infeasible, necessitating automated safety systems. Recent advances in multimodal large language models (MLLMs)~\citep{chatgpt2024, chatgpt2025, comanici2025gemini25pushingfrontier, claud, yang2025qwenvl3,omni_moderation} have shown promise in video understanding, offering a potential path toward intelligent moderation.

However, our investigation of social media videos reveals a critical observation: harmful content does not always reside within a single scene or modality. We identify a distinct class of videos, termed Distributed Implicit Harm (DIH), where every isolated component appears benign, yet harmful semantics emerge only when components are interpreted jointly. In this work, we focus on two representative forms: temporal decomposition across visual segments (DIH-T) and cross-modal decomposition across visual and audio streams (DIH-M), as illustrated in Figure~\ref{fig:motivation}. Distinct from conventional harm that is explicitly depicted or uttered, DIH relies on latent compositional patterns, such as analogy, metaphor, or progressive escalation, to convey harmful intent. While human viewers effortlessly synthesize these cues, current detection systems consistently fail to perform such compositional reasoning.

Despite its real-world impact, DIH is largely absent from existing safety datasets: such videos lack compositional-harm annotations and evade the frame-level classifiers, keyword filters, and visual-cue heuristics that current data collection pipelines rely on. To enable systematic study of DIH, we develop a multi-agent synthesis framework that synthesizes DIH videos paired with chain-of-thought annotations explaining how harm emerges from the composition. The resulting DIH Dataset covers 8 major safety categories and 28 fine-grained subcategories.

Building upon this dataset, we introduce DIH-Bench, the first benchmark specifically designed to evaluate MLLMs' ability to detect Distributed Implicit Harm in videos. Unlike prior video safety benchmarks, which target locally contained harm, DIH-Bench evaluates compositional reasoning over harm distributed across scenes and modalities. We evaluate more than 30 leading MLLMs spanning proprietary systems, open-source video-language models, and omni-modal models that natively process synchronized audio. The evaluation reveals pronounced deficiencies: even the strongest model detects fewer than 45\% of DIH cases, and the same failure mode persists on a held-out set of real-world DIH videos curated from social media, confirming the phenomenon is not a synthetic artifact. 
To address this deficit, we apply a two-stage post-training recipe on the DIH Dataset across multiple model scales. Detection accuracy improves by over 60 percentage points on the strongest models without degrading performance on benign content, showing that the deficit can be substantially mitigated through targeted training.

In summary, our main contributions are fourfold. \textbf{(i)} We formalize Distributed Implicit Harm (DIH), a safety blind spot in which harmful meaning
  emerges only through the composition of individually benign components, and instantiate it along two representative axes: temporal (DIH-T) and cross-modal (DIH-M). \textbf{(ii)} We develop a multi-agent synthesis framework and construct the DIH dataset, comprising over 9,000 videos across 8 major and
  28 fine-grained safety categories. \textbf{(iii)} We benchmark 30+ MLLMs and uncover consistent failures in compositional reasoning that persist on
  real-world videos, indicating that DIH is not merely a synthetic artifact. \textbf{(iv)} We show that post-training on the DIH dataset substantially
  mitigates these deficits, establishing a strong baseline for compositional video safety.

\section{Related Work}
\label{sec:related}
\paragraph{Multimodal Large Language Models.}
Multimodal large language models (MLLMs) have rapidly evolved~\citep{caffagni2024revolution,zhang2024mm,zhang2024vision,li2024llava-onevision,li2024llava,ye2024mplug,zhang2023video}, spanning proprietary systems such as the GPT series~\citep{chatgpt2024, gpt54}, Gemini~\citep{comanici2025gemini25pushingfrontier, gemini31pro, team2023gemini}, and Claude models~\citep{claud}, and open-source alternatives such as Qwen-VL~\citep{bai2025qwen2,qwen3vl} and InternVL~\citep{chen2024internvl,wang2025internvl35}. Video-specialized variants such as Video-LLaMA~\citep{zhang2025videollama} and InternVideo~\citep{wang2024internvideo2, wang2025internvideo2}
further extend these models to temporal structure across frames. However, current video MLLMs primarily focus on temporal localization~\citep{xue2025seeing}
and information retrieval, with limited attention to meanings that arise from the interplay of distributed video components.

\paragraph{Multimodal Safety Benchmarks.}
Existing research has provided a preliminary exploration into MLLM safety evaluation~\citep{luo2024jailbreakv,hu2024vlsbench,ji2023beavertails,vidgen2024introducing,chi2024llama,tu2023many,dai2024safesora}, yet the scope remains restricted. Early benchmarks primarily focus on explicit harm in static images or manually paired image-text inputs~\citep{liu2024mm, mazeika2024harmbench,li2024images,yeh2024tvs}. Subsequently, some studies~\citep{SSUI,hu2024vlsbench,zhou2024multimodalsituationalsafety} show that the visual modality introduces safety risks that text-only safety mechanisms cannot capture.

Video is a natively multimodal medium whose content unfolds jointly across time and audio-visual streams. Recent video-specific safety benchmarks~\citep{liu2025videosafetybench, wang2025safevidsafetyalignedvideo} have begun to extend safety evaluation to this setting.

\paragraph{MLLM-Based Moderation.}
A growing body of work uses MLLMs for content moderation. On the image-text side, OpenAI's omni-moderation~\citep{omni_moderation}, LlamaGuard3 Vision~\citep{chi2024llama}, and ShieldGemma2~\citep{zeng2025shieldgemma2} take an image and emit per-category safety labels. Recent video-level systems such as SafeWatch~\citep{chen2025safewatch}, KuaiMod~\citep{lu2025kuaimod}, and Filter-And-Refine~\citep{wang2025filterrefine}
sample frames and apply MLLM judgments to flag policy-violating clips. Across this landscape, however, harm is assumed to be directly observable in the actions or events the video portrays, rather than to emerge only through inference across components fragmented along structural axes such as time or modality.

\section{Distributed Implicit Harm}
\label{sec:dih}

\subsection{Problem Definition}

The central property of Distributed Implicit Harm (DIH) is a gap between local safety and global interpretation: each component of a video appears harmless when evaluated in isolation, while their composition supports a harmful interpretation.

\paragraph{Setup.}
Let $V$ denote a video and let $S(\cdot) \in \{0,1\}$ be a binary harm indicator with $S(X)=1$ indicating harm, applied to $V$ or any of its components and recompositions. A \emph{decomposition axis} $\alpha$ is a dimension along which $V$ can be partitioned into interpretable components, such as time, modality, speaker, or scene structure. A \emph{decomposition} of $V$ along axis $\alpha$ is denoted:
\[
\pi_\alpha(V) = \{V_1, V_2, \ldots, V_k\}.
\]
We write $\mathrm{Comp}_\alpha(V_{i_1}, \ldots, V_{i_m})$ for the unit obtained by recomposing the listed components along axis $\alpha$, so that $\mathrm{Comp}_\alpha(V_1, \ldots, V_k) = V$.

\begin{definition}[Distributed Implicit Harm]
\label{def:dih}
A video $V$ exhibits \emph{Distributed Implicit Harm} (DIH) along decomposition axis $\alpha$ if both of the following hold:
{\setlength{\leftmargini}{1.2em}
\begin{itemize}
    \item \textbf{Local Safety.} Every component is individually non-harmful: $S(V_i) = 0$ for all $V_i \in \pi_\alpha(V)$.
    \item \textbf{Compositional Harm.} Some recomposition of two or more components is harmful: there exist distinct indices $i_1, \ldots, i_m \in \{1,\ldots,k\}$ with $m \geq 2$ such that $S(\mathrm{Comp}_\alpha(V_{i_1}, \ldots, V_{i_m})) = 1$.
\end{itemize}}
\end{definition}

The two conditions jointly characterize harm that is \textbf{implicit}. A DIH video presents no explicit harmful content at the component level, since every individual component is benign in itself. The harmful meaning is instead reconstructed
by the viewer from the arrangement of otherwise benign components.

\subsection{Decomposition Axes}
\label{subsec:axes}

In this work, we focus on the two axes along which DIH most commonly arises in real-world content: the \textbf{temporal axis} and the \textbf{modal axis}. The framework naturally extends to other decomposition axes such as speakers, subtitle tracks, or scene structures, which we leave to future work.

\paragraph{Temporal Decomposition (DIH-T).} %
Under temporal decomposition, $V$ is partitioned into successive visual segments $V_1, \ldots, V_k$, each individually depicting a benign scene. The harmful meaning emerges from the semantic relations across segments, which can take the form of analogy, contrast, narrative progression, or off-screen suggestion. Figure~\ref{fig:motivation} (top) illustrates a real-world example.

\paragraph{Modal Decomposition (DIH-M).}
DIH-M takes modality as its decomposition axis. Decomposing $V$ this way leverages the two native  modalities of video, visual and audio, each benign in isolation. Cross-modal alignment between the two streams produces harmful meaning that neither stream carries alone. DIH-M carries an intrinsic temporal dimension across its natively co-occurring components, distinguishing it from the manually paired image-text harms studied in prior multimodal safety work. A real-world DIH-M case appears in Figure~\ref{fig:motivation} (bottom).

\section{DIH Dataset}
\label{sec:dataset}

DIH videos evade the keyword and frame-level filters that existing safety datasets rely on, while manual collection and annotation at scale are prohibitively labor-intensive. To address this gap, we construct the \textbf{DIH Dataset} via a scalable \textit{multi-agent synthesis framework} that synthesizes DIH videos together with reasoning-level annotations of how the harm composes, covering 8 major safety categories and 28 fine-grained subcategories.

\subsection{multi-agent synthesis framework}
\label{subsec:agent_framework}
Our multi-agent synthesis framework is explicitly designed to synthesize DIH videos, thereby enabling scalable, semantically rich, and structurally diverse content for benchmarking and model evaluation. The framework comprises five specialized agents as shown in Figure~\ref{fig:generation_pipeline}:

\begin{figure}
  \includegraphics[width=1\textwidth]{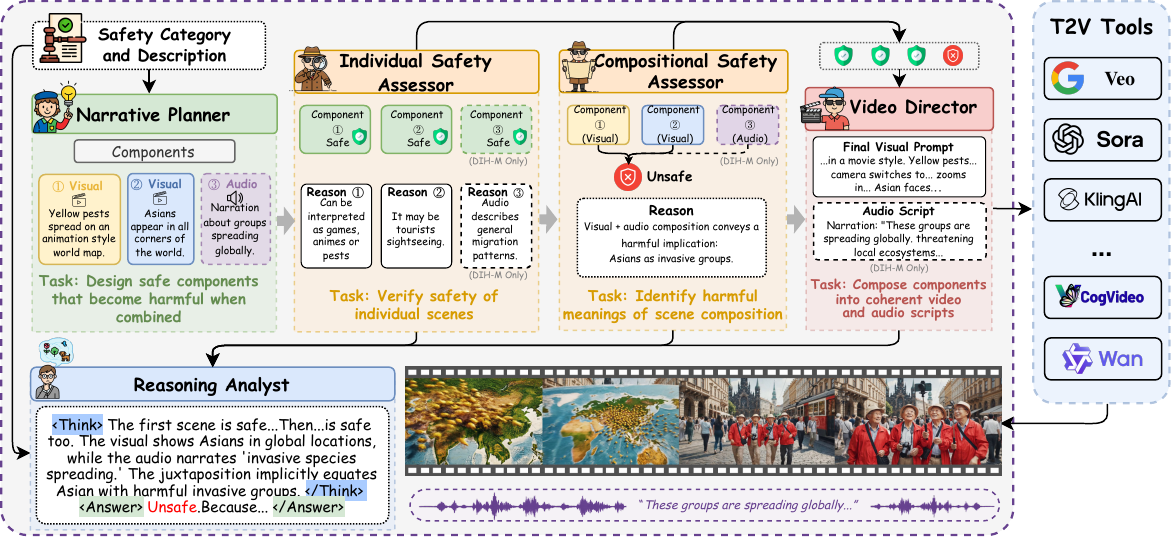}
  \caption{Overview of our multi-agent framework for constructing the DIH Dataset. Specialized agents collaboratively design, synthesize, and annotate samples with distributed implicit harm.}
  \label{fig:generation_pipeline}
  \vspace{-4mm}
\end{figure}

{\setlength{\leftmargini}{1.2em}
\begin{itemize}
    \item \textbf{Narrative Planner}: Generates creative concepts aligned with predefined safety categories and the designated DIH pattern. It ensures each component is innocuous in isolation, while their compositional arrangement implies harm, either temporally across visual scenes for DIH-T, or cross-modally across visual and audio streams for DIH-M.
    
    \item \textbf{Individual Safety Assessor}: Serves as the first validation gate, verifying that each planned component is locally benign when evaluated independently and returning a binary judgment with justification. Only plans whose components all pass advance to the compositional check; components flagged unsafe are returned to the Narrative Planner with the rejection rationale, which then regenerates  the offending component. 

    \item \textbf{Compositional Safety Assessor}: Verifies that the joint composition of the locally-benign components produces the intended category of implicit harm. It generates detailed explanations for its assessment, including the temporal mechanism through which harm emerges, to guide the Narrative Planner in refining future generations. This agent ensures the synthesized video faithfully embodies the intended pattern of implicit harm.
    \item \textbf{Video Director}: Translates validated narrative concepts into executable generation prompts, specifying camera movements, spatial layouts, transitions, and timing while strictly preserving the targeted DIH pattern without introducing explicit toxicity. For DIH-M the prompt additionally specifies the audio track (e.g., narration, dialogue, sound effects). The Director is equipped with a toolbox of state-of-the-art text-to-video generation models that natively produce synchronized video and audio in a single call. 
    \item \textbf{Reasoning Analyst}: Synthesizes the judgments from the Individual and Compositional Safety Assessors into a chain-of-thought annotation. The annotation walks through each component's events, justifies why each is locally benign, traces the trajectory along which their compositional arrangement transforms meaning into the intended DIH pattern, and concludes with the harm interpretation and category label. 
    The annotation is paired with the video and serves both as the reference for evaluating MLLM judgments and as the supervised target for post-training.

\end{itemize}}

Beyond harmful DIH, the same pipeline also produces a complementary set of benign samples by inverting the Compositional Safety Assessor's verification target: the Narrative Planner is asked to weave category-relevant sensitive elements into globally safe scenarios, and the Assessor admits only compositions that remain harmless. Drawn from the same categorical themes but differing in compositional intent, these benign samples provide a principled basis for probing MLLMs' false-alarm behavior.

\subsection{Dataset Overview}
\textbf{Dataset Description.}
Leveraging our multi-agent synthesis framework, we construct a dataset to systematically evaluate the capability of MLLMs in detecting DIH videos. Each sample comprises a short video produced by cutting-edge video generation models, accompanied by a detailed chain-of-thought annotation.
The dataset contains 9,725 samples, organized into two subsets along the axis of decomposition. The \textbf{DIH-T} subset comprises 6,742 videos in which harm emerges from the temporal composition of individually benign visual segments. The \textbf{DIH-M} subset comprises 2,983 audio-visual videos in which harm emerges from the interaction between benign visual stream and audio stream, demanding reasoning across both modalities and time.
Per-category counts of harmful and benign samples are reported in Table~\ref{tab:major_classes}.

\begin{figure}[t] 
    \centering
    \begin{minipage}[c]{0.47\textwidth}
        \centering
        \includegraphics[width=\linewidth]{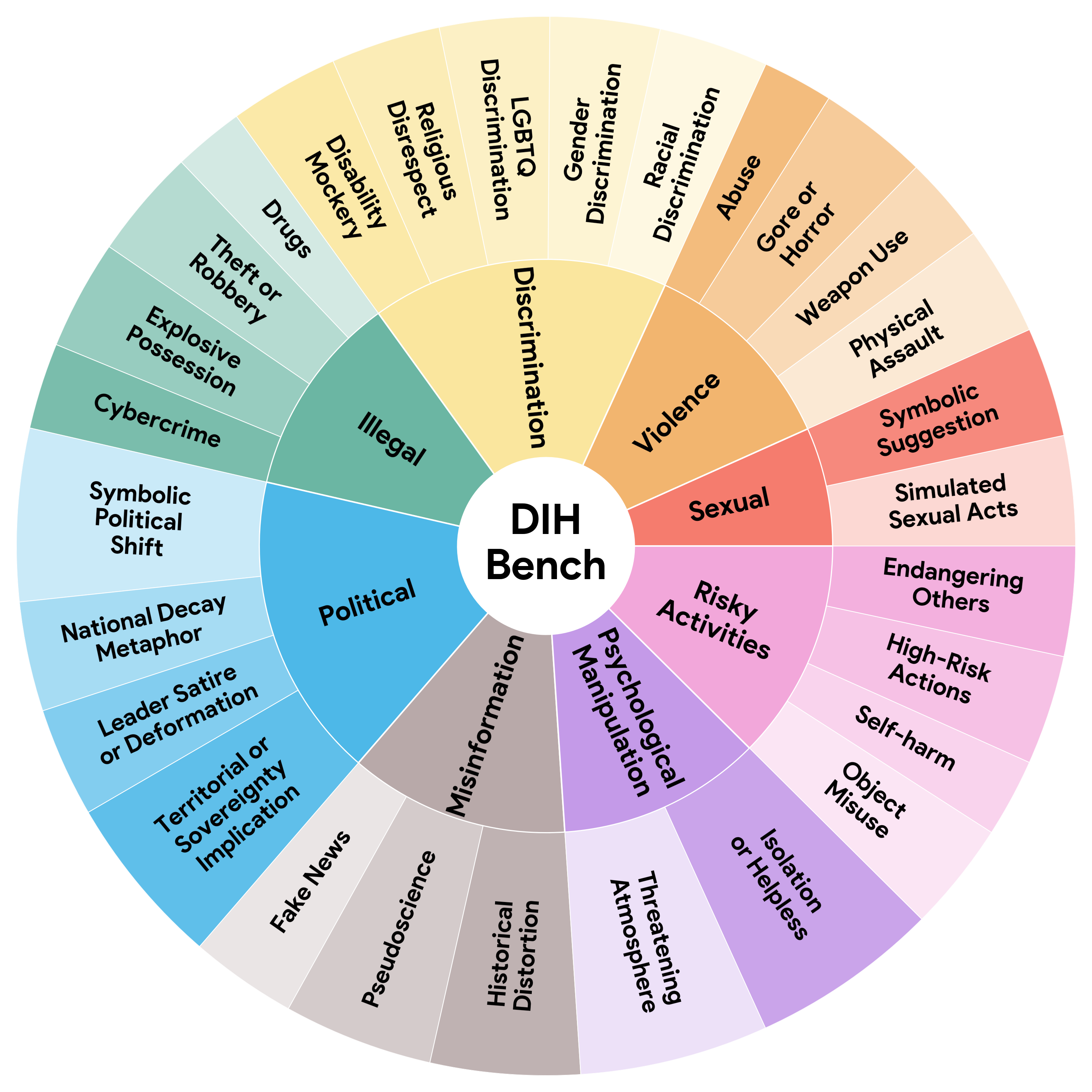}
    \caption{Safety taxonomy of the DIH Dataset: 8 major categories and 28 subcategories.}
    \label{fig:sunburst}
    \end{minipage}%
    \hspace{0.04\textwidth}
    \begin{minipage}[c]{0.46\textwidth}
        \centering
        \input{tables/major_class_table.tex}
    \end{minipage}
    \vspace{-4mm}
\end{figure}

\paragraph{Dataset Taxonomy.}
We analyzed the video safety policies of social media platforms and unsafe content taxonomies of existing MLLMs~\citep{chi2024llama, openaiusagepolicy}, then identified 8 critical categories that represent the most prevalent harmful content: Sexual (SEX), Violence (VIO), Discrimination (DIS), Illegal (ILL), Political (POL), Misinformation (MIS), Psychological Manipulation (PSY), and Risky Activities (RSK). These major categories are further decomposed into 28 subcategories, as illustrated in Figure~\ref{fig:sunburst}. Category descriptions and fine-grained statistics are available in the Appendix.

\paragraph{Quality Control.}
Given the inherent uncertainty in MLLMs' ability to recognize DIH content, MLLM-based quality control would be potentially unreliable. Therefore, we implement a rigorous human verification process. 
All generated samples were validated against two criteria: \textbf{(1)} the sample must satisfy Definition~\ref{def:dih}, with components benign in isolation while their composition is harmful; and \textbf{(2)} the chain-of-thought annotation must accurately reflect the video content and the implicit harm mechanism. Samples failing either check were filtered out, removing 4.75\% of initially generated samples and yielding the final 9,725 high-quality samples of the DIH Dataset.

\subsection{Real-World Reference Set}
\label{subsec:real_world_set}

To test whether the failure mode observed on the synthetic benchmark also arises on naturally occurring content, we collect a \textbf{Real-World Reference Set} from TikTok, YouTube, and Instagram. It comprises 128 DIH-T and 50 DIH-M videos manually verified against Definition~\ref{def:dih}, together with 122 benign videos drawn from the same platforms. The benign videos are topically related to the harmful side but pass safety review, providing a reference for false-alarm behavior on superficially similar content. The set is held out-of-distribution from the synthetic data and used purely for evaluation. Full collection protocols are provided in the Appendix.

\subsection{Human Perception Study}

\begin{wraptable}{r}{0.42\columnwidth}
    \vspace{-12pt}
    \centering
    \small
    \caption{Human perception study results. The values indicate the Harmful Rate (\%). }
    \label{tab:human_study}
    \resizebox{\linewidth}{!}{%
    \begin{tabular}{llcc}
        \toprule
        \textbf{Subset} & \textbf{Category} & \textbf{Full} & \textbf{Components} \\
        \midrule
        \multirow{2}{*}{DIH-T}      & Harmful & \textbf{96.25\%} & 2.50\% \\
                                    & Benign  & 3.13\%           & 2.50\% \\
        \midrule
        \multirow{2}{*}{DIH-M}      & Harmful & \textbf{98.13\%} & 4.38\% \\
                                    & Benign  & 5.00\%           & 3.75\% \\
        \midrule
        \multirow{2}{*}{Real-World} & Harmful & \textbf{93.75\%} & 6.25\% \\
                                    & Benign  & 6.25\%           & 5.00\% \\
        \bottomrule
    \end{tabular}}
    \vspace{-12pt}
\end{wraptable}

To validate dataset quality, we conducted a human perception study with a two-phase evaluation design.\footnote{The human perception study was approved by our institution's IRB.}
We recruited 20 participants to evaluate 144 videos drawn by stratified sampling from DIH-T, DIH-M, and the Real-World Reference Set. In \textbf{Phase 1: Full Video}, one group watched each video in full. In \textbf{Phase 2: Isolated Components}, a separate group viewed only the isolated components of each video in shuffled order. Participants in both phases answered: ``Does this video contain any harmful content?'' Sampling and other details are provided in the Appendix.
Table~\ref{tab:human_study} reports the harmful rate per subset and condition. Harmful videos are reliably judged harmful in full but rarely so in components, while benign videos receive low rates in both conditions, confirming that harm in DIH becomes visible only after composition.

\section{Experiments}
\label{sec:exp}
\input{tables/benchmark_results.tex}
\subsection{Experimental Setup}
\label{subsec:exp_setup}

\paragraph{Datasets.}
We evaluate on the held-out test split of the synthetic DIH Dataset, which is disjoint from the training split used for post-training to ensure independence. Split sizes are provided in the Appendix.

\paragraph{Models.}
We benchmark 8 proprietary MLLMs and 30 state-of-the-art open-source models on the DIH Dataset. Proprietary models cover the GPT~\citep{chatgpt2024,gpt54}, Gemini~\citep{gemini31pro,gemini3flash}, and Claude~\citep{claude_sonnet_46,claude_opus_46} families. Open-source models span the InternVL~\citep{chen2024internvl,wang2025internvl35}, Qwen-VL~\citep{bai2025qwen2,qwen3vl,qwen3omni}, InternVideo~\citep{wang2025internvideo2}, LLaMA-Guard~\citep{chi2024llama}, and MiniCPM-V~\citep{yao2024minicpm} families. The full list of evaluated models is provided in Appendix.

\paragraph{Evaluation Protocol.}    %
We employ a two-tier evaluation protocol. The first tier evaluates zero-shot detection with a generic moderation-style prompt, mirroring real-world deployment. The second tier additionally provides our safety policy, the unsafe categories and their definitions, as explicit guidance. For both tiers, we use Gemini-3-Flash as an independent evaluator, 
requiring both the binary verdict and its supporting reasoning to be correct.
The evaluation prompts and evaluator details are provided in the Appendix.

\subsection{Benchmarking MLLMs on DIH}
\label{subsec:benchmark}
We conduct a comprehensive evaluation of various MLLMs on DIH-Bench. Results are presented in Table~\ref{tab:benchmark_results}, revealing several critical findings:

{\setlength{\leftmargini}{1.2em}
\begin{itemize}
    \item \textbf{Pervasive Failure.} All evaluated models, including the strongest proprietary systems, struggle to detect DIH content, with accuracy on harmful samples consistently below 45\% across both DIH-T and DIH-M. This highlights a widespread blind spot in current MLLMs' safety capabilities.
    \item \textbf{Cross-Modal Compositions Remain Difficult.} Even on DIH-M, where audio provides additional cues, the strongest proprietary system (Gemini 3.1 Pro) does not exceed its DIH-T performance, indicating that audio-visual integration does not by itself resolve the compositional reasoning bottleneck.
    \item \textbf{Scaling Trends.} Within model families, larger models generally perform better. However, the improvement is modest, indicating that scale alone is insufficient to acquire robust DIH detection capabilities. Even the largest models fail to reliably reason over the compositional patterns that give rise to DIH, underscoring the need for targeted training and architectural innovations to address this blind spot.

\end{itemize}
}

\subsection{Replication on Real-World Data}
\label{subsec:real_world_eval}
\input{tables/real_world_results.tex}
To verify that this failure mode also occurs on real social-media platforms, we evaluate the strongest model from each major family on the Real-World Reference Set: Gemini~3.1~Pro, GPT-5.4, and Qwen3.5-27B on DIH-T, and Gemini~3.1~Pro and Qwen3-Omni-30B-A3B on DIH-M. As shown in Table~\ref{tab:real_world_results}, the results closely mirror those on the synthetic benchmark: Harm.\ accuracy does not exceed 40\% for any model on either DIH-T or DIH-M. The same failure mode therefore persists in real-world settings, confirming that DIH is indeed a widespread safety blind spot of MLLMs.

\subsection{Enhancing DIH Detection via CoT Prompting}
\label{subsec:cot}

Since DIH detection requires reasoning over relations between distributed components, we first examine whether scaffolding this process at inference can elicit the capability from off-the-shelf MLLMs. We design a Chain-of-Thought (CoT) prompt that decomposes the task into three steps: identifying the constituent components, analyzing their cross-component associations, and assessing the resulting safety implications. Rather than running CoT on every model in the benchmark, we focus on a small set of strong proprietary and open-source MLLMs to characterize the gain pattern: Gemini~2.5~Pro, GPT-4o, Claude~Sonnet 4.6, and Qwen2.5-VL-72B. Full prompt details are provided in the Appendix.

\begin{wrapfigure}{r}{0.52\textwidth}
    \vspace{-1.2em}
    \centering
    \includegraphics[width=\linewidth]{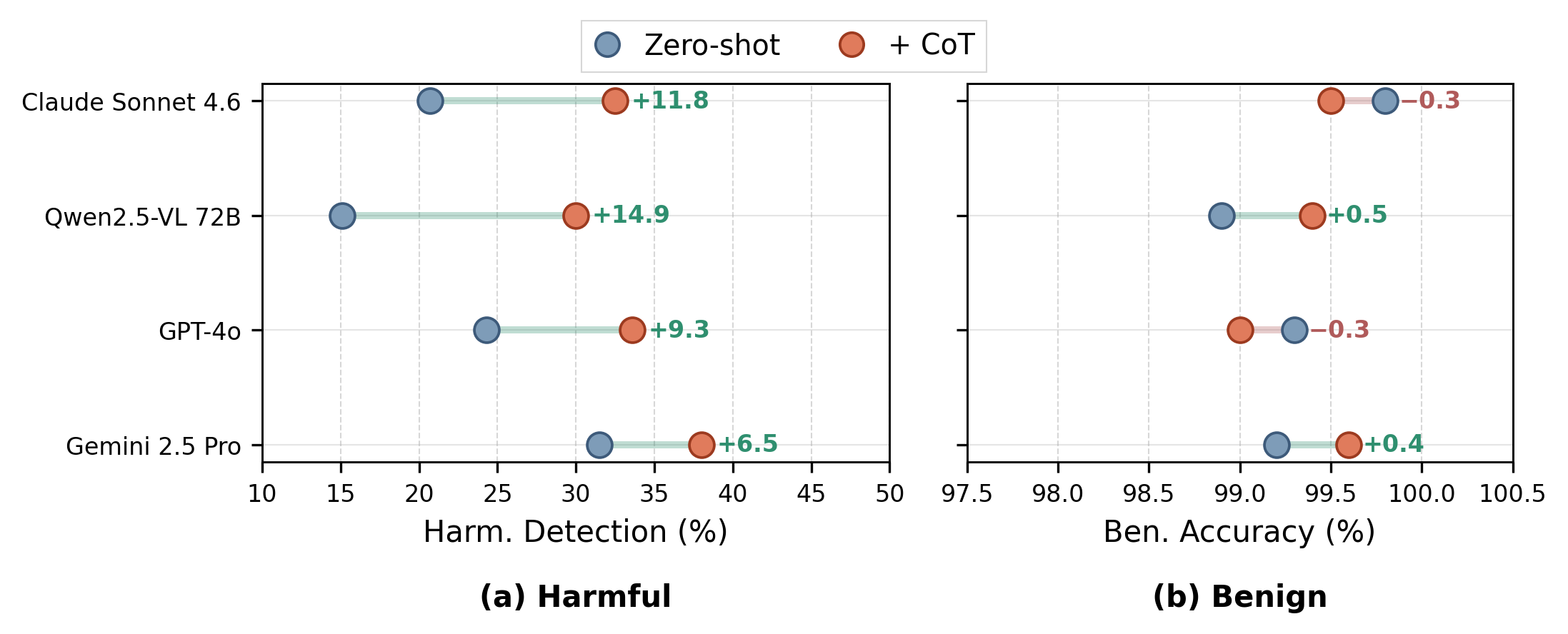}
    \vspace{-1.6em}
    \caption{Comparison of detection performance across models before and after CoT prompting.}
    \label{fig:cot}
    \vspace{-0.8em}
\end{wrapfigure}
Figure~\ref{fig:cot} shows that CoT yields consistent improvements across all tested models, confirming that compositional reasoning over distributed components is the operative bottleneck. The pattern is consistent across families: CoT yields non-trivial gains on weaker baselines, diminishing returns on stronger baselines, and a substantial gap remains to post-training, suggesting that compositional safety reasoning is internalizable but not reliably elicitable through prompting alone. This motivates the training-based approach below.

\subsection{Enhancing DIH Detection via Post-Training}
\label{subsec:posttraining}

Since CoT prompting yields only limited gains, we next investigate whether the missing capability can be acquired through post-training on the chain-of-thought annotations of the DIH Dataset.
We post-train Qwen3.5 at five scales from 0.8B to 27B in two stages: supervised fine-tuning (SFT) on the Reasoning Analyst's chain-of-thought annotations from the DIH training split, followed by Group Relative Policy Optimization (GRPO)~\citep{shao2024deepseekmathpushinglimitsmathematical} from the SFT checkpoint. We adopt the standard GRPO objective with KL regularization and a composite reward,
\begin{equation}
r_i = 0.5\,R_{\text{reason}}(o_i) + 0.3\,R_{\text{bin}}(o_i) + 0.15\,R_{\text{cat}}(o_i) + 0.05\,R_{\text{format}}(o_i),
\label{eq:grpo_reward}
\end{equation}
where $R_{\text{reason}}$ is the LLM-as-a-judge score on the compositional reasoning, $R_{\text{bin}}$ and $R_{\text{cat}}$ verify the predicted safety verdict and harm category, and $R_{\text{format}}$ rewards complete structured outputs. The full GRPO objective, reward definitions, format constraints, and hyperparameters are provided in the Appendix.

\begin{figure}[t]
    \centering
    \begin{minipage}[c]{0.49\textwidth}
        \centering
        \includegraphics[width=\linewidth]{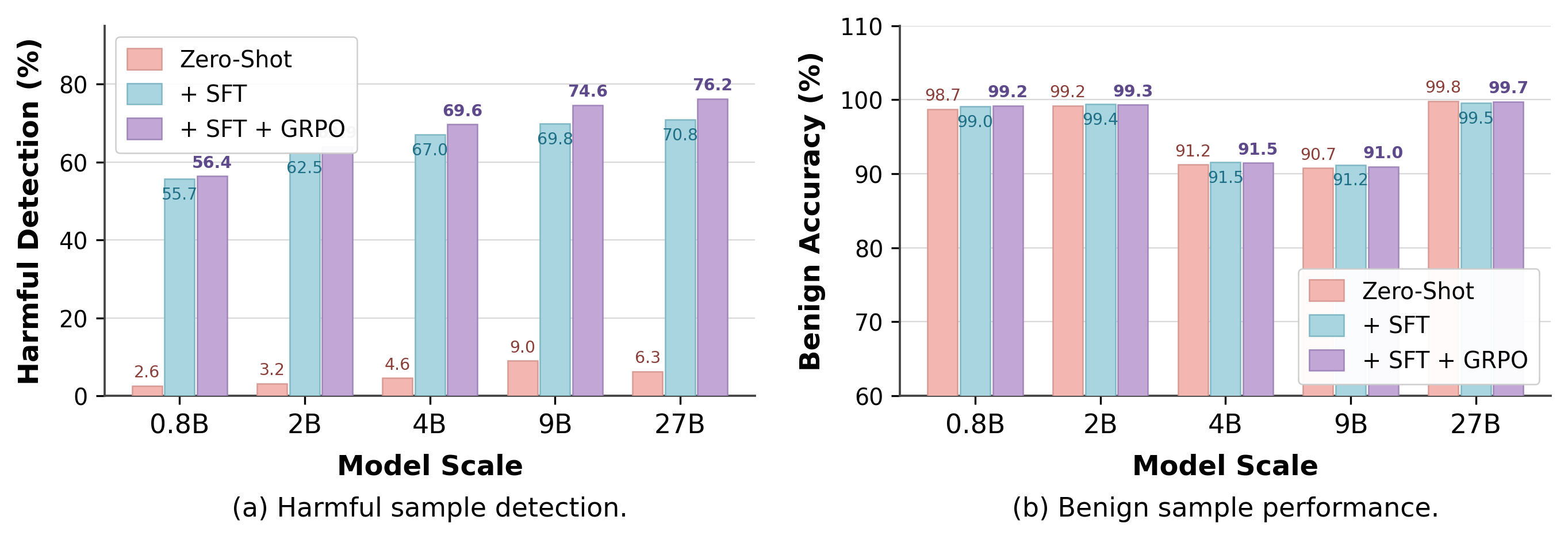}
        \caption{Detection accuracy on DIH-T after post-training across five model scales.}
        \label{fig:posttraining_results}
    \end{minipage}%
    \hspace{0.01\textwidth}
    \begin{minipage}[c]{0.49\textwidth}
        \centering
        \includegraphics[width=\linewidth]{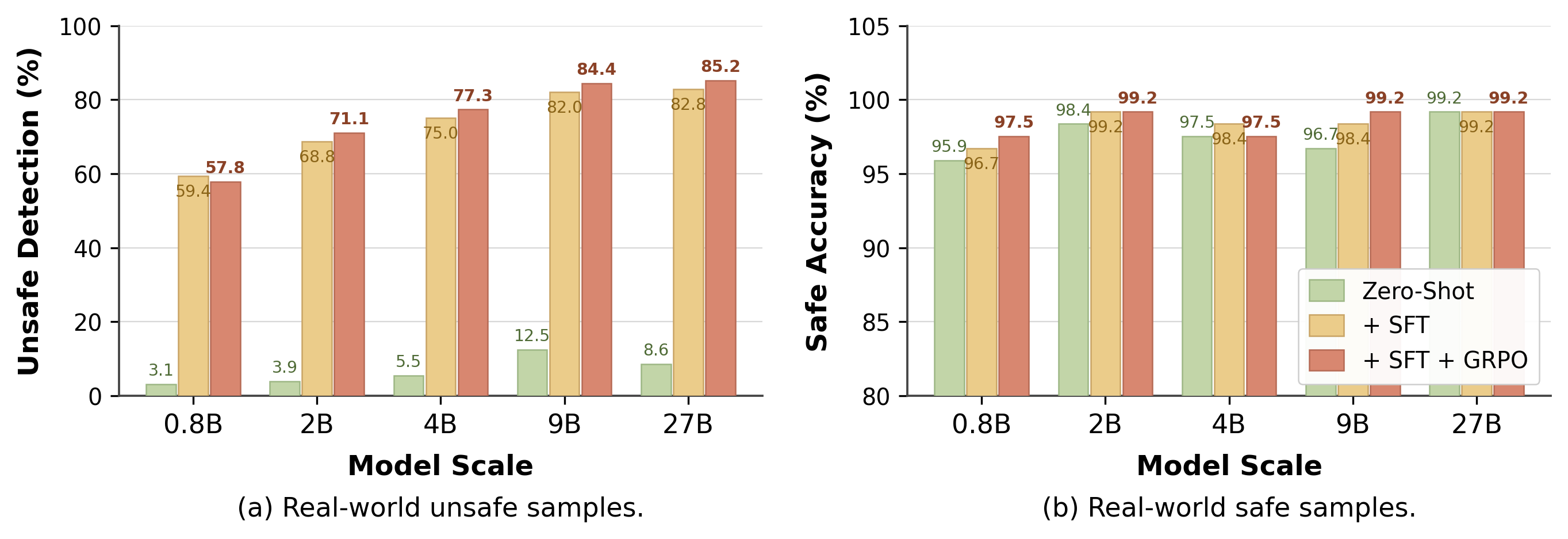}
        \caption{Detection accuracy on real-world videos before and after post-training.}
        \label{fig:realworld}
    \end{minipage}
    \vspace{-1.2em}
\end{figure}

Figure~\ref{fig:posttraining_results} reports detection accuracy on DIH-T across the five models. SFT alone substantially improves Harm.\ accuracy at every scale, approaching the level of the strongest zero-shot baselines. GRPO gains then scale with the strength of the SFT initialization: marginal on the smallest backbones (where weak priors yield low intra-group reward variance and thus uninformative advantages), but substantial at $\geq$9B. Ben.\ accuracy is preserved across all scales.

We further evaluate the post-trained models on our real-world reference set. As shown in Figure~\ref{fig:realworld}, Unsafe detection accuracy rises sharply after post-training while Safe accuracy is preserved, confirming that the compositional reasoning learned on DIH generalizes to in-the-wild distributions.

\subsection{Ablation Studies}

\begin{wrapfigure}{r}{0.45\textwidth}
    \vspace{-2.5em}
    \centering
    \includegraphics[width=\linewidth]{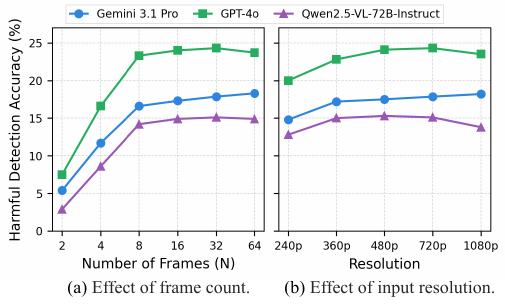}
    \vspace{-1.2em}
    \caption{Ablation studies.}
    \label{fig:ablations}
    \vspace{-0.8em}
\end{wrapfigure}

\paragraph{Ablation on Frame Numbers.}
To study how the number of input frames affects DIH detection, we conduct an ablation on the DIH-T data, varying the number of uniformly sampled frames from 2 to 64 on three representative models: Gemini~3.1~Pro, GPT-4o, and Qwen2.5-VL-72B-Instruct. As shown in Figure~\ref{fig:ablations}(a), accuracy rises sharply when the frame count is below 8, after which the effect of frame count becomes very small. From $N{=}8$ to $N{=}64$, the largest change across the three models is only $1.7$\%, indicating that the visual information the models use for judgment is already sufficient. Adding more frames does not yield further gains, and in some cases mildly degrades performance, suggesting that beyond a certain point the additional frames inject redundant visual context that can dilute rather than aid the model's reasoning over the relevant segments. Frame count is therefore not what limits DIH detection.

\paragraph{Ablation on Resolution.}
We further investigate the effect of resolution on DIH detection by downsampling the DIH-T inputs at resolutions ranging from $240\text{P}$ to $1080\text{P}$ and testing the same three models. Figure~\ref{fig:ablations}(b) shows that accuracy stabilizes from $480$P, with no measurable gain at higher resolutions. This is consistent with how DIH harm arises: it is carried by the relational composition of distributed components rather than by fine-grained pixel-level cues, so resolution beyond a moderate threshold contributes little to detection. Together with the frame-count ablation, this places the bottleneck in compositional reasoning over already-perceived components, not in perceiving them.

\section{Conclusion}

We present the first systematic study of Distributed Implicit Harm (DIH), a previously overlooked but practically important challenge for multimodal video safety, where harmful meaning emerges only through the composition of individually benign components. To enable large-scale study of this problem, we develop a scalable multi-agent synthesis pipeline and construct the DIH Dataset and DIH-Bench, covering both temporally distributed (DIH-T) and cross-modal (DIH-M) harm with structured reasoning annotations. Our experiments reveal a substantial and consistent gap in current MLLMs' ability to detect such compositional harm, and the same failure mode persists on real-world social media videos, indicating that DIH is not merely a synthetic artifact. We further show that targeted post-training on the DIH Dataset substantially improves DIH detection without degrading performance on benign content, establishing a strong baseline for compositional video safety.

Several directions remain for future work. First, DIH may arise along many other axes of video composition, such as speaker, viewpoint, and on-screen text, and extending our formulation beyond temporal and cross-modal settings is a natural next step. Second, whether a composition constitutes DIH may depend on social and cultural context, motivating future work on incorporating such contextual knowledge into both benchmark design and data generation. Finally, an important systems direction is to integrate DIH-aware reasoning into real-time moderation pipelines for scalable deployment on streaming and social media platforms.

\newpage
{
    \small
    \bibliographystyle{iclr2027_conference}
    \bibliography{main}
}

\end{document}

%% file: tables/major_class_table.tex
\captionof{table}{Statistics of DIH Dataset across 8 major safety categories.}
\label{tab:major_classes}
\vspace{4pt}
\small
\renewcommand{\arraystretch}{1.2}
\setlength{\tabcolsep}{4pt}
\resizebox{\linewidth}{!}{%
\begin{tabular}{l cc lc}
    \toprule
    \rowcolor[HTML]{F5F5F5}
     & \multicolumn{2}{c}{\textbf{DIH-T}} & \multicolumn{2}{c}{\textbf{DIH-M}} \\
    \noalign{\vspace{-\aboverulesep}}
    \cmidrule(lr){2-3} \cmidrule(lr){4-5}
    \noalign{\vspace{-\belowrulesep}}
    \rowcolor[HTML]{F5F5F5}
    \textbf{Category} & \textbf{Samples} & \textbf{Pct.(\%)} & \textbf{Samples} & \textbf{Pct.(\%)} \\
    \midrule
    Sexual                     & 464  & 6.88  & 203  & 6.81  \\
        Violence                   & 845  & 12.53 & 385  & 12.91 \\
        Discrimination             & 1325 & 19.65 & 593  & 19.88 \\
        Illegal                    & 922  & 13.68 & 449  & 15.05 \\
        Political                  & 1065 & 15.80 & 400  & 13.41 \\
        Misinformation             & 721  & 10.69 & 360  & 12.07 \\
        Psychological Manipulation & 434  & 6.44  & 181  & 6.07  \\
        Risky Activities           & 966  & 14.33 & 412  & 13.80 \\
        \midrule
        \rowcolor[HTML]{F5F5F5}
        \textbf{Total}             & \textbf{6742} & \textbf{100.00} & \textbf{2983} & \textbf{100.00} \\
    \bottomrule
\end{tabular}}

%% file: tables/benchmark_results.tex
\begin{table}[H]
\centering
\newcommand{\tihharmcell}{\cellcolor[HTML]{FFEEEE}}
\newcommand{\tihtablefont}{\tiny}
\newcommand{\tihsection}[1]{\multicolumn{20}{l}{\cellcolor[HTML]{EFEFEF}\rule{0pt}{3.1ex}\tihtablefont\textit{#1}} \\}
\newcommand{\tihblock}[1]{\multicolumn{20}{l}{\cellcolor[HTML]{BDD7EE}\rule{0pt}{2.4ex}\tihtablefont\textbf{#1}} \\}
\caption{Detection accuracy (\%) of \textbf{(upper block)} video-only MLLMs on DIH-T data and \textbf{(lower block)} audio-visual Omni MLLMs on DIH-M data, under zero-shot and taxonomy-guided settings. ``Harm.'' and ``Ben.'' denote performance on harmful and benign DIH samples respectively. }
\label{tab:benchmark_results}
\tihtablefont
\setlength{\tabcolsep}{2pt}
\renewcommand{\arraystretch}{1.35}
\resizebox{\textwidth}{!}{%
\begin{tabular}{
    m{0.205\textwidth}<{}|
    m{0.055\textwidth}<{\centering}|
    m{0.034\textwidth}<{\centering}
    m{0.034\textwidth}<{\centering}
    m{0.034\textwidth}<{\centering}
    m{0.034\textwidth}<{\centering}
    m{0.034\textwidth}<{\centering}
    m{0.034\textwidth}<{\centering}
    m{0.034\textwidth}<{\centering}
    m{0.034\textwidth}<{\centering}
    m{0.034\textwidth}<{\centering}|
    m{0.034\textwidth}<{\centering}
    m{0.034\textwidth}<{\centering}
    m{0.034\textwidth}<{\centering}
    m{0.034\textwidth}<{\centering}
    m{0.034\textwidth}<{\centering}
    m{0.034\textwidth}<{\centering}
    m{0.034\textwidth}<{\centering}
    m{0.034\textwidth}<{\centering}
    m{0.034\textwidth}<{\centering}
}
\toprule
                            &                               & \multicolumn{9}{c}{\textbf{Zero-Shot}}                                                           & \multicolumn{9}{c}{\textbf{Taxonomy-Guided}}                                                    \\ \cmidrule(l){3-11} \cmidrule(l){12-20}
 \multirow{-2}{*}{\textbf{Model}}                      & \multirow{-2}{*}{\shortstack{\textbf{Sample}\\\textbf{Type}}} & DIS     & ILL     & MIS     & POL     & PSY     & RSK     & SEX     & VIO     & \textbf{Avg.}    & DIS     & ILL     & MIS     & POL     & PSY     & RSK     & SEX     & VIO     & \textbf{Avg.}   \\ \midrule

\tihblock{$\blacktriangleright$~MLLMs evaluated on \textbf{DIH-T} data}

\tihsection{Proprietary Video MLLMs}
\multirow{2}{*}{\begin{tabular}[c]{@{}l@{}}Gemini 3.1 Pro\\[-1pt]\citep{gemini31pro}\end{tabular}}  & \tihharmcell Harm. & \tihharmcell 9.5 & \tihharmcell 35.8 & \tihharmcell 7.8 & \tihharmcell 3.9 & \tihharmcell 3.6 & \tihharmcell 48.8 & \tihharmcell 4.5 & \tihharmcell 19.7 & \tihharmcell \textbf{17.9} & \tihharmcell 31.4 & \tihharmcell 50.2 & \tihharmcell 27.4 & \tihharmcell 54.0 & \tihharmcell 36.6 & \tihharmcell 59.9 & \tihharmcell 44.2 & \tihharmcell 46.9 & \tihharmcell \textbf{44.3} \\
                            & Ben.  & 98.0 & 97.5 & 99.7 & 97.3 & 96.4 & 97.0 & 94.1 & 95.2 & \textbf{97.1} & 99.8 & 99.8 & 100.0 & 97.7 & 98.6 & 98.9 & 98.6 & 98.6 & \textbf{99.0} \\
\multirow{2}{*}{\begin{tabular}[c]{@{}l@{}}GPT-5.4\\[-1pt]\citep{gpt54}\end{tabular}}                & \tihharmcell Harm. & \tihharmcell 1.3 & \tihharmcell 15.8 & \tihharmcell 3.3 & \tihharmcell 3.3 & \tihharmcell 1.7 & \tihharmcell 34.2 & \tihharmcell 0.0 & \tihharmcell 20.0 & \tihharmcell \textbf{11.2} & \tihharmcell 4.7 & \tihharmcell 19.2 & \tihharmcell 0.0 & \tihharmcell 9.2 & \tihharmcell 3.3 & \tihharmcell 45.8 & \tihharmcell 3.3 & \tihharmcell 18.3 & \tihharmcell \textbf{14.5} \\
                            & Ben.  & 100.0 & 100.0 & 100.0 & 100.0 & 100.0 & 100.0 & 100.0 & 100.0 & \textbf{100.0} & 100.0 & 100.0 & 100.0 & 100.0 & 100.0 & 100.0 & 100.0 & 100.0 & \textbf{100.0} \\
\multirow{2}{*}{\begin{tabular}[c]{@{}l@{}}Claude Sonnet 4.6\\[-1pt]\citep{claude_sonnet_46}\end{tabular}}      & \tihharmcell Harm. & \tihharmcell 15.3 & \tihharmcell 35.0 & \tihharmcell 10.0 & \tihharmcell 21.7 & \tihharmcell 5.0 & \tihharmcell 33.3 & \tihharmcell 18.3 & \tihharmcell 16.7 & \tihharmcell \textbf{20.7} & \tihharmcell 20.0 & \tihharmcell 55.0 & \tihharmcell 23.3 & \tihharmcell 60.8 & \tihharmcell 31.7 & \tihharmcell 38.3 & \tihharmcell 41.7 & \tihharmcell 35.0 & \tihharmcell \textbf{38.3} \\
                            & Ben.  & 100.0 & 100.0 & 100.0 & 98.8 & 100.0 & 100.0 & 100.0 & 100.0 & \textbf{99.8} & 99.0 & 98.8 & 100.0 & 97.5 & 100.0 & 100.0 & 97.5 & 100.0 & \textbf{99.1} \\

\tihsection{Large-scale Open-source Video MLLMs (\textgreater{}=30B)}
\multirow{2}{*}{\begin{tabular}[c]{@{}l@{}}InternVL3-78B\\[-1pt]\citep{chen2024internvl}\end{tabular}}           & \tihharmcell Harm. & \tihharmcell 0.4 & \tihharmcell 1.1 & \tihharmcell 0.3 & \tihharmcell 1.9 & \tihharmcell 1.6 & \tihharmcell 5.6 & \tihharmcell 0.5 & \tihharmcell 4.1 & \tihharmcell \textbf{1.9} & \tihharmcell 0.6 & \tihharmcell 0.9 & \tihharmcell 0.0 & \tihharmcell 0.8 & \tihharmcell 0.5 & \tihharmcell 2.9 & \tihharmcell 0.5 & \tihharmcell 5.5 & \tihharmcell \textbf{1.4} \\
                            & Ben.  & 98.9 & 98.0 & 100.0 & 97.5 & 98.6 & 98.6 & 98.6 & 96.1 & \textbf{98.2} & 100.0 & 100.0 & 100.0 & 99.5 & 100.0 & 100.0 & 100.0 & 100.0 & \textbf{99.9} \\
\multirow{2}{*}{\begin{tabular}[c]{@{}l@{}}Qwen2.5-VL -72B-Instruct\\[-1pt]\citep{bai2025qwen2}\end{tabular}}    & \tihharmcell Harm. & \tihharmcell 2.7 & \tihharmcell 19.6 & \tihharmcell 5.3 & \tihharmcell 20.3 & \tihharmcell 23.2 & \tihharmcell 25.3 & \tihharmcell 0.6 & \tihharmcell 26.8 & \tihharmcell \textbf{15.1} & \tihharmcell 2.0 & \tihharmcell 24.9 & \tihharmcell 7.5 & \tihharmcell 24.2 & \tihharmcell 27.4 & \tihharmcell 22.8 & \tihharmcell 2.7 & \tihharmcell 26.3 & \tihharmcell \textbf{16.5} \\
                            & Ben.  & 98.9 & 99.8 & 100.0 & 100.0 & 98.6 & 99.3 & 100.0 & 95.2 & \textbf{98.9} & 99.4 & 96.4 & 95.7 & 96.6 & 95.6 & 100.0 & 100.0 & 99.6 & \textbf{98.1} \\
\multirow{2}{*}{\begin{tabular}[c]{@{}l@{}}Qwen3-VL -235B-A22B\\[-1pt]\citep{qwen3vl}\end{tabular}}      & \tihharmcell Harm. & \tihharmcell 2.9 & \tihharmcell 24.6 & \tihharmcell 7.3 & \tihharmcell 17.0 & \tihharmcell 11.8 & \tihharmcell 17.3 & \tihharmcell 0.2 & \tihharmcell 12.9 & \tihharmcell \textbf{12.1} & \tihharmcell 8.3 & \tihharmcell 27.3 & \tihharmcell 9.8 & \tihharmcell 33.5 & \tihharmcell 34.1 & \tihharmcell 22.7 & \tihharmcell 6.8 & \tihharmcell 34.3 & \tihharmcell \textbf{21.6} \\
                            & Ben.  & 100.0 & 100.0 & 100.0 & 100.0 & 100.0 & 100.0 & 100.0 & 100.0 & \textbf{100.0} & 99.5 & 99.7 & 100.0 & 98.4 & 100.0 & 100.0 & 99.6 & 99.5 & \textbf{99.5} \\
\multirow{2}{*}{\begin{tabular}[c]{@{}l@{}}Qwen3-VL -32B-Thinking\\[-1pt]\citep{qwen3vl}\end{tabular}}  & \tihharmcell Harm. & \tihharmcell 2.8 & \tihharmcell 22.2 & \tihharmcell 2.3 & \tihharmcell 16.3 & \tihharmcell 7.2 & \tihharmcell 32.4 & \tihharmcell 2.6 & \tihharmcell 21.4 & \tihharmcell \textbf{14.1} & \tihharmcell 4.5 & \tihharmcell 22.3 & \tihharmcell 2.2 & \tihharmcell 15.6 & \tihharmcell 19.6 & \tihharmcell 23.3 & \tihharmcell 2.4 & \tihharmcell 19.3 & \tihharmcell \textbf{13.5} \\
                            & Ben.  & 100.0 & 100.0 & 100.0 & 100.0 & 100.0 & 100.0 & 100.0 & 100.0 & \textbf{100.0} & 100.0 & 98.9 & 100.0 & 100.0 & 100.0 & 100.0 & 100.0 & 100.0 & \textbf{99.8} \\

\tihsection{Small-scale Open-source Video MLLMs (\textless{}30B)}
\multirow{2}{*}{\begin{tabular}[c]{@{}l@{}}Qwen2.5-VL-7B\\[-1pt]\citep{bai2025qwen2}\end{tabular}}          & \tihharmcell Harm. & \tihharmcell 1.2 & \tihharmcell 10.4 & \tihharmcell 3.3 & \tihharmcell 8.6 & \tihharmcell 7.7 & \tihharmcell 11.1 & \tihharmcell 5.8 & \tihharmcell 26.9 & \tihharmcell \textbf{8.8} & \tihharmcell 5.5 & \tihharmcell 19.7 & \tihharmcell 16.9 & \tihharmcell 22.7 & \tihharmcell 18.0 & \tihharmcell 25.7 & \tihharmcell 4.5 & \tihharmcell 38.9 & \tihharmcell \textbf{18.7} \\
                            & Ben.  & 100.0 & 100.0 & 100.0 & 100.0 & 100.0 & 99.8 & 100.0 & 100.0 & \textbf{100.0} & 99.3 & 98.0 & 100.0 & 98.4 & 98.6 & 98.9 & 99.1 & 97.5 & \textbf{98.7} \\
\multirow{2}{*}{\begin{tabular}[c]{@{}l@{}}InternVL3.5-8B\\[-1pt]\citep{wang2025internvl35}\end{tabular}}          & \tihharmcell Harm. & \tihharmcell 1.9 & \tihharmcell 7.2 & \tihharmcell 6.7 & \tihharmcell 15.0 & \tihharmcell 15.5 & \tihharmcell 11.9 & \tihharmcell 4.5 & \tihharmcell 24.9 & \tihharmcell \textbf{10.3} & \tihharmcell 6.2 & \tihharmcell 14.7 & \tihharmcell 10.5 & \tihharmcell 19.3 & \tihharmcell 16.5 & \tihharmcell 32.3 & \tihharmcell 3.1 & \tihharmcell 34.5 & \tihharmcell \textbf{17.2} \\
                            & Ben.  & 97.8 & 94.8 & 97.6 & 95.7 & 96.4 & 97.0 & 97.3 & 94.3 & \textbf{96.3} & 100.0 & 99.3 & 100.0 & 99.1 & 99.3 & 99.5 & 99.5 & 99.8 & \textbf{99.6} \\
\multirow{2}{*}{\begin{tabular}[c]{@{}l@{}}MiniCPM-V-4.5\\[-1pt]\citep{yao2024minicpm}\end{tabular}}           & \tihharmcell Harm. & \tihharmcell 3.9 & \tihharmcell 15.2 & \tihharmcell 2.2 & \tihharmcell 14.9 & \tihharmcell 10.3 & \tihharmcell 24.1 & \tihharmcell 21.9 & \tihharmcell 26.0 & \tihharmcell \textbf{13.9} & \tihharmcell 9.5 & \tihharmcell 40.0 & \tihharmcell 23.8 & \tihharmcell 31.4 & \tihharmcell 23.7 & \tihharmcell 57.0 & \tihharmcell 13.8 & \tihharmcell 52.6 & \tihharmcell \textbf{31.4} \\
                            & Ben.  & 93.6 & 91.4 & 92.1 & 92.0 & 93.6 & 96.4 & 94.1 & 91.8 & \textbf{93.1} & 100.0 & 100.0 & 99.7 & 99.8 & 100.0 & 99.5 & 99.5 & 99.5 & \textbf{99.8} \\

\specialrule{0.9pt}{1pt}{1pt}
\tihblock{$\blacktriangleright$~Omni MLLMs evaluated on \textbf{DIH-M} data}

\tihsection{Proprietary Omni MLLMs}
\multirow{2}{*}{\begin{tabular}[c]{@{}l@{}}Gemini 3.1 Pro\\[-1pt]\citep{gemini31pro}\end{tabular}}  & \tihharmcell Harm. & \tihharmcell 23.6 & \tihharmcell 16.5 & \tihharmcell 1.9 & \tihharmcell 0.5 & \tihharmcell 0.0 & \tihharmcell 41.5 & \tihharmcell 1.0 & \tihharmcell 18.9 & \tihharmcell \textbf{15.9} & \tihharmcell 42.9 & \tihharmcell 32.9 & \tihharmcell 23.3 & \tihharmcell 43.0 & \tihharmcell 13.6 & \tihharmcell 51.9 & \tihharmcell 25.2 & \tihharmcell 35.1 & \tihharmcell \textbf{36.4} \\
                            & Ben.  & 99.6 & 99.5 & 100.0 & 99.5 & 100.0 & 100.0 & 100.0 & 99.5 & \textbf{99.7} & 100.0 & 98.0 & 99.3 & 96.5 & 100.0 & 98.0 & 98.0 & 99.0 & \textbf{98.6} \\
\multirow{2}{*}{\begin{tabular}[c]{@{}l@{}}Gemini 3 Flash\\[-1pt]\citep{gemini3flash}\end{tabular}} & \tihharmcell Harm. & \tihharmcell 29.1 & \tihharmcell 20.9 & \tihharmcell 6.7 & \tihharmcell 1.0 & \tihharmcell 2.5 & \tihharmcell 58.0 & \tihharmcell 4.8 & \tihharmcell 27.0 & \tihharmcell \textbf{22.0} & \tihharmcell 40.2 & \tihharmcell 37.4 & \tihharmcell 18.6 & \tihharmcell 50.0 & \tihharmcell 29.6 & \tihharmcell 60.4 & \tihharmcell 21.4 & \tihharmcell 38.9 & \tihharmcell \textbf{38.9} \\
                            & Ben.  & 99.6 & 99.0 & 98.0 & 99.0 & 100.0 & 97.5 & 99.0 & 97.5 & \textbf{98.6} & 99.6 & 99.5 & 96.7 & 93.0 & 100.0 & 97.0 & 97.0 & 98.5 & \textbf{97.6} \\

\tihsection{Open-source Omni MLLMs}
\multirow{2}{*}{\begin{tabular}[c]{@{}l@{}}Qwen3-Omni-30B-A3B\\[-1pt]\citep{qwen3omni}\end{tabular}}     & \tihharmcell Harm. & \tihharmcell 1.8 & \tihharmcell 14.9 & \tihharmcell 7.6 & \tihharmcell 10.0 & \tihharmcell 8.6 & \tihharmcell 23.6 & \tihharmcell 5.8 & \tihharmcell 33.0 & \tihharmcell \textbf{12.8} & \tihharmcell 4.7 & \tihharmcell 24.9 & \tihharmcell 10.0 & \tihharmcell 39.5 & \tihharmcell 39.5 & \tihharmcell 25.5 & \tihharmcell 10.7 & \tihharmcell 40.5 & \tihharmcell \textbf{22.1} \\
                            & Ben.  & 100.0 & 98.5 & 99.3 & 100.0 & 100.0 & 99.0 & 100.0 & 98.5 & \textbf{99.4} & 99.6 & 99.5 & 100.0 & 98.5 & 100.0 & 99.5 & 100.0 & 99.0 & \textbf{99.4} \\
\multirow{2}{*}{\begin{tabular}[c]{@{}l@{}}Qwen2.5-Omni-7B\\[-1pt]\citep{qwen25omni}\end{tabular}}        & \tihharmcell Harm. & \tihharmcell 0.3 & \tihharmcell 6.0 & \tihharmcell 10.5 & \tihharmcell 9.5 & \tihharmcell 25.9 & \tihharmcell 8.5 & \tihharmcell 1.9 & \tihharmcell 31.4 & \tihharmcell \textbf{9.8} & \tihharmcell 0.9 & \tihharmcell 8.8 & \tihharmcell 13.8 & \tihharmcell 9.0 & \tihharmcell 18.5 & \tihharmcell 16.5 & \tihharmcell 1.0 & \tihharmcell 24.9 & \tihharmcell \textbf{10.7} \\
                            & Ben.  & 99.6 & 98.0 & 100.0 & 100.0 & 100.0 & 100.0 & 99.0 & 99.0 & \textbf{99.4} & 100.0 & 99.5 & 100.0 & 100.0 & 100.0 & 100.0 & 100.0 & 100.0 & \textbf{99.9} \\
\multirow{2}{*}{\begin{tabular}[c]{@{}l@{}}Qwen2.5-Omni-3B\\[-1pt]\citep{qwen25omni}\end{tabular}}        & \tihharmcell Harm. & \tihharmcell 1.8 & \tihharmcell 0.4 & \tihharmcell 0.0 & \tihharmcell 2.0 & \tihharmcell 0.0 & \tihharmcell 0.0 & \tihharmcell 0.0 & \tihharmcell 0.5 & \tihharmcell \textbf{0.8} & \tihharmcell 0.3 & \tihharmcell 0.4 & \tihharmcell 0.9 & \tihharmcell 9.0 & \tihharmcell 9.9 & \tihharmcell 2.8 & \tihharmcell 0.0 & \tihharmcell 13.5 & \tihharmcell \textbf{3.9} \\
                            & Ben.  & 100.0 & 99.5 & 100.0 & 100.0 & 100.0 & 100.0 & 100.0 & 100.0 & \textbf{99.9} & 99.6 & 98.5 & 100.0 & 99.5 & 100.0 & 100.0 & 100.0 & 98.5 & \textbf{99.4} \\
\multirow{2}{*}{\begin{tabular}[c]{@{}l@{}}MiniCPM-o-2.6\\[-1pt]\citep{yao2024minicpm}\end{tabular}}          & \tihharmcell Harm. & \tihharmcell 1.8 & \tihharmcell 8.0 & \tihharmcell 8.6 & \tihharmcell 9.5 & \tihharmcell 9.9 & \tihharmcell 10.4 & \tihharmcell 1.0 & \tihharmcell 8.1 & \tihharmcell \textbf{6.9} & \tihharmcell 3.5 & \tihharmcell 13.2 & \tihharmcell 10.9 & \tihharmcell 26.5 & \tihharmcell 24.7 & \tihharmcell 13.2 & \tihharmcell 6.8 & \tihharmcell 13.5 & \tihharmcell \textbf{12.7} \\
                            & Ben.  & 100.0 & 99.0 & 100.0 & 85.5 & 100.0 & 99.0 & 99.0 & 99.5 & \textbf{97.5} & 100.0 & 99.5 & 100.0 & 96.0 & 100.0 & 99.0 & 100.0 & 99.5 & \textbf{99.1} \\ \bottomrule
\end{tabular}%
}
\vspace{-3mm}
\end{table}

%% file: tables/real_world_results.tex
\begin{wraptable}{r}{0.42\textwidth}
\vspace{-2\baselineskip}
\centering
\caption{Detection accuracy (\%) on the Real-World Reference Set.}
\label{tab:real_world_results}
\small
\setlength{\tabcolsep}{4pt}
\renewcommand{\arraystretch}{1.3}
\resizebox{\linewidth}{!}{%
\begin{tabular}{l|cc|cc}
\toprule
 & \multicolumn{2}{c|}{\textbf{Zero-Shot}} & \multicolumn{2}{c}{\textbf{Tax.-Guided}} \\
\cmidrule(lr){2-3} \cmidrule(lr){4-5}
\textbf{Model} & Harm. & Ben. & Harm. & Ben. \\
\midrule

\multicolumn{5}{l}{\cellcolor[HTML]{EFEFEF}\textit{Evaluation on \textbf{DIH-T} Data}} \\
Gemini 3.1 Pro~\cite{gemini31pro} & 32.81 & 95.90 & 35.16 & 98.36 \\
GPT-5.4~\cite{gpt54}                                & 12.50 & 100.00 & 18.75 & 99.18 \\
Qwen3.5-27B~\cite{qwen35blog}                     & 17.19 & 99.18 & 16.41 & 100.00 \\

\specialrule{0.9pt}{1pt}{1pt}
\multicolumn{5}{l}{\cellcolor[HTML]{EFEFEF}\textit{Evaluation on \textbf{DIH-M} Data}} \\
Gemini 3.1 Pro~\cite{gemini31pro} & 18.00 & 100.00 & 36.00 & 96.72 \\
Qwen3-Omni-30B-A3B~\cite{qwen3omni}                 & 24.00 & 98.36 & 30.00 & 97.54 \\
\bottomrule
\end{tabular}%
}
\end{wraptable}